%% file: main.tex
\documentclass[conference]{IEEEtran}
\IEEEoverridecommandlockouts

\usepackage{times}  
\usepackage{helvet}  
\usepackage{courier}  
\usepackage[hyphens]{url}  
\usepackage{comment}

\usepackage[square,sort,comma,numbers]{natbib}

\usepackage{caption} 

\usepackage[hyphens]{url}
\usepackage[]{hyperref}
\usepackage{tikz,tikz-qtree}
\usepackage{booktabs}
\usepackage{multirow}

\hypersetup{breaklinks=true}
\usepackage{cite}
\usepackage{amsmath,amssymb,amsfonts}
\usepackage{algorithmic}
\usepackage{graphicx}
\usepackage{textcomp}

\usepackage{algorithm}
\usepackage{algorithmic}
\usepackage{amsmath}
\usepackage{amssymb}
\usepackage{amsthm}
\usepackage{url}
\usepackage{empheq}
\usepackage{bbm}

\theoremstyle{definition}
\newtheorem{definition}{Definition}

\newtheoremstyle{subdefinition}
  {0pt}       
  {0pt}       
  {\upshape}  
  {0pt}       
  {\bfseries} 
  {.}         
  {5pt plus 1pt minus 1pt} 
  {}          
\theoremstyle{subdefinition}
\newtheorem{subdefinition}{Definition}[definition]

\newtheorem{theorem}{Theorem}
\newtheorem{lemma}[subdefinition]{Lemma}

\usepackage{amsmath}               
  {
      \theoremstyle{plain}
      
  }

\theoremstyle{remark}

\usepackage{newfloat}
\usepackage{listings}
\DeclareCaptionStyle{ruled}{labelfont=normalfont,labelsep=colon,strut=off} 
\floatstyle{ruled}
\newfloat{listing}{tb}{lst}{}
\floatname{listing}{Listing}

\def\BibTeX{{\rm B\kern-.05em{\sc i\kern-.025em b}\kern-.08em
    T\kern-.1667em\lower.7ex\hbox{E}\kern-.125emX}}

\def\ninept{\def\baselinestretch{.95}\let\normalsize\small\normalsize}

\begin{document}

\title{Towards Unbiased Evaluation of Time-series Anomaly Detector \\
\thanks{This work was carried out as part of the IBM Research Global Internship Program 2024. }
}

\author{
\IEEEauthorblockN{Debarpan Bhattacharya\IEEEauthorrefmark{1}, Sumanta Mukherjee\IEEEauthorrefmark{2}, Chandramouli Kamanchi\IEEEauthorrefmark{2}, Vijay Ekambaram\IEEEauthorrefmark{2}, Arindam Jati\IEEEauthorrefmark{2},\\ Pankaj Dayama\IEEEauthorrefmark{2}}
\IEEEauthorblockA{\IEEEauthorrefmark{1}\textit{Indian Institute of Science}, Bangalore, India \\
}
\IEEEauthorblockA{\IEEEauthorrefmark{2}\textit{IBM Research}, Bangalore, India \\
Email: pankajdayama@in.ibm.com
}
}
\ninept
\maketitle

\setlength{\abovedisplayskip}{3pt}
\setlength{\belowdisplayskip}{3pt}

\input{sections/abstract}
\input{sections/introduction}

\input{sections/related_works}
\input{sections/methods}
\input{sections/experiments}

\input{sections/conclusions}

\bibliographystyle{IEEEbib}
\bibliography{mybib}

\end{document}

%% file: sections/abstract.tex
\begin{abstract}
Time series anomaly detection (TSAD) is an evolving area of research motivated by its critical applications, such as detecting seismic activity, sensor failures in industrial plants, predicting crashes in the stock market, and so on. Across domains, anomalies occur significantly less frequently than normal data, making the F1-score the most commonly adopted metric for anomaly detection. However, in the case of time series, it is not straightforward to use standard F1-score because of the dissociation between `time points' and `time events'. To accommodate this, anomaly predictions are adjusted, called as point adjustment (PA), before the $F_1$-score evaluation. However, these adjustments are heuristics-based, and biased towards true positive detection, resulting in over-estimated detector performance. In this work, we propose an alternative adjustment protocol called ``Balanced point adjustment'' (BA). It addresses the limitations of existing point adjustment methods and provides guarantees of fairness backed by axiomatic definitions of TSAD evaluation. Code and implementation details: \url{https://github.com/summukhe/balanced_f1score}.
\end{abstract}

\begin{IEEEkeywords}
time series, anomaly detection, point adjustment, F1 score.
\vspace{-0.05in}
\end{IEEEkeywords}

%% file: sections/introduction.tex
\section{Introduction}

Anomaly detection plays a crucial role in identifying system failures or performance deviations. Anomalies are rare data patterns, often detected by comparing the likelihood of an instance with respect to the background distribution. As a result, anomaly and outlier detection are closely related, with applications spanning tabular, image, and audio data. The detection process involves classifying observations as either anomalous or normal, making it similar to binary classification. Consequently, binary classification metrics, such as the receiver operating characteristics (ROC)\citep{hanley1982meaning} and \emph{F1 score}\citep{rijsbergen1979information}, are commonly used for Time Series Anomaly Detection (TSAD). ROC captures detector behavior across thresholds, while the $F_1$ score is crucial for assessing a detector’s performance, especially in the setting of imbalanced data with low anomaly ratios~\citep{sarkar2024can}.

An anomaly detection system typically has two components: the scorer and the detector~\citep{anomaly-survey}. The scorer processes signals into a score, and the detector determines a threshold for detecting anomalies. The ROC curve informs threshold selection, while the \emph{F1 score} evaluates the detector's overall performance.

Applying the standard $F_1$-score to time series presents challenges due to the contiguous nature of anomalies. Point adjustment (PA), introduced by~\citep{xu2018unsupervised}, modifies predictions before evaluating the $F_1$ score to account for this. However, PA often biases results in favor of true positives, leading to inflated performance estimates~\citep{pa-ka}. While alternatives like $F_{1KPA}$ exist, their advantages over $F_{1PA}$ remain unclear.

\begin{figure}
\centering 
\includegraphics[width=\linewidth]{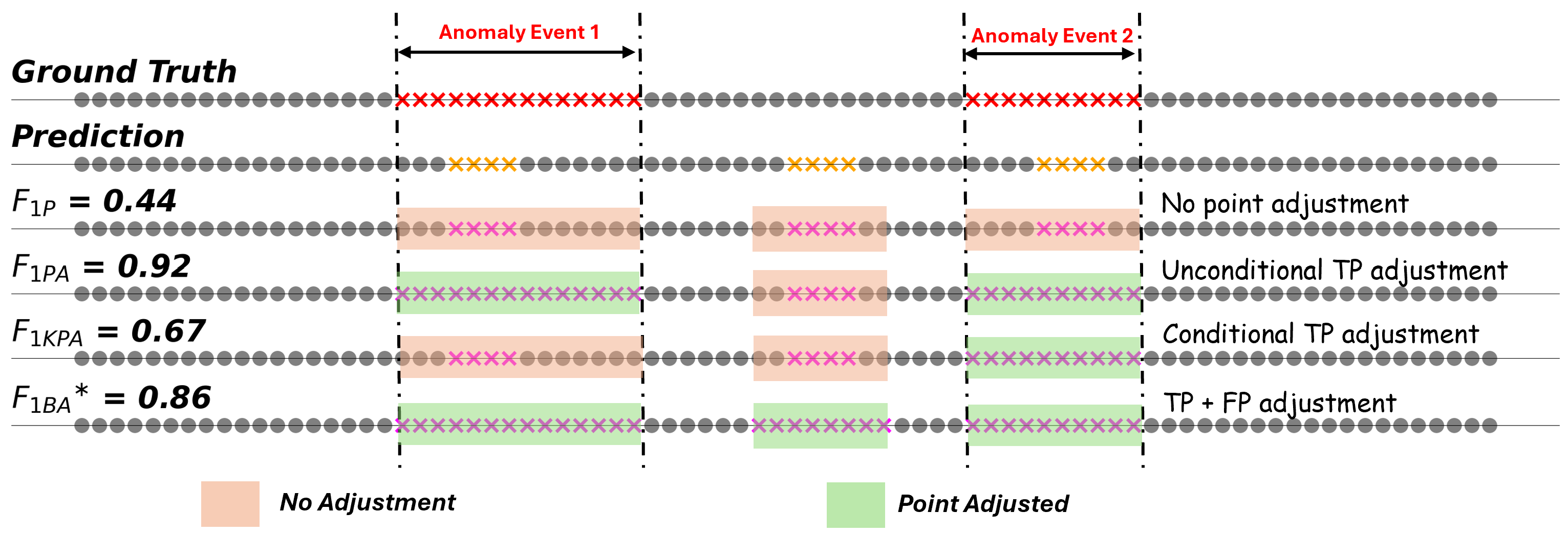}
\caption{A comparative view of different point adjustment methods for a given ground truth and predicted labels. We have computed the $F_{1KPA}$ score with $K=40\%$. $F_{1BA}$ is the proposed method in this paper. $F_{1BA}$ is the only metric that penalizes false positive detection. The \textcolor{orange}{orange} highlights detection which is left as it is, and \textcolor{green}{green} highlights describe instances that are adjusted before $F_{1}$ score computation.}
\label{fig:overview}
\vspace{-0.2in}
\end{figure}

We introduce a new protocol, `Balanced Point Adjustment' (BA), to address these biases. BA penalizes false positives and balances the adjustments made for true positives, providing a more accurate and fair evaluation of TSAD models. This method, as shown in Figure~\ref{fig:overview}, ensures a more reliable assessment of anomaly detectors, supported by controlled experiments.

%% file: sections/related_works.tex
\section{Related Works} Time series anomalies often span multiple time points, and partial detection is typically considered valid~\citep{range-anomaly-neurips}. To accommodate this, the \emph{F1 score} for time series anomaly detection (TSAD) is computed after point adjustment. Point adjustment extends a detection across the anomaly span, which is then used in the \emph{F1 score} calculation~\citep{xu2018unsupervised}. However, this introduces bias by overestimating model accuracy, as even random scores can produce favorable results~\citep{metric-taxonomy,pa-ka}. \citep{pa-ka} proposed a stricter adjustment based on a $k\%$ threshold, though its improvements over the original point adjustment remain unclear.

There is ambiguity in the field regarding the appropriate choice of the metric for TSAD due to the shortcomings of pointwise $F_1$ ($F_{1p}$), biased point-adjusted $F_1$ ($F_{1PA}$), and heuristic-based corrections like $F_{1KPA}$. \begin{itemize} \item Many studies still use $F_{1PA}$ despite its limitations~\citep{xu2021anomaly, chen2021learning, audibert2020usad,zhou2022contrastive,zhang2022tfad,wu2023decompose}, even for evaluating state-of-the-art models~\citep{zhou2023one, goswamimoment}. \item Some studies have adopted $F_{1KPA}$ for TSAD evaluation~\citep{choi2024self, garg2021evaluation}. \item A few revert to using pointwise $F_1$ ($F_{1p}$)\citep{darban2024carla,zamanzadeh2022deep,li2022robust,yao2024svd}, while others combine all three metrics\citep{choi2024self}. \end{itemize}

\begin{figure}[t!] \centering \includegraphics[width=8.7cm, height=4.8cm]{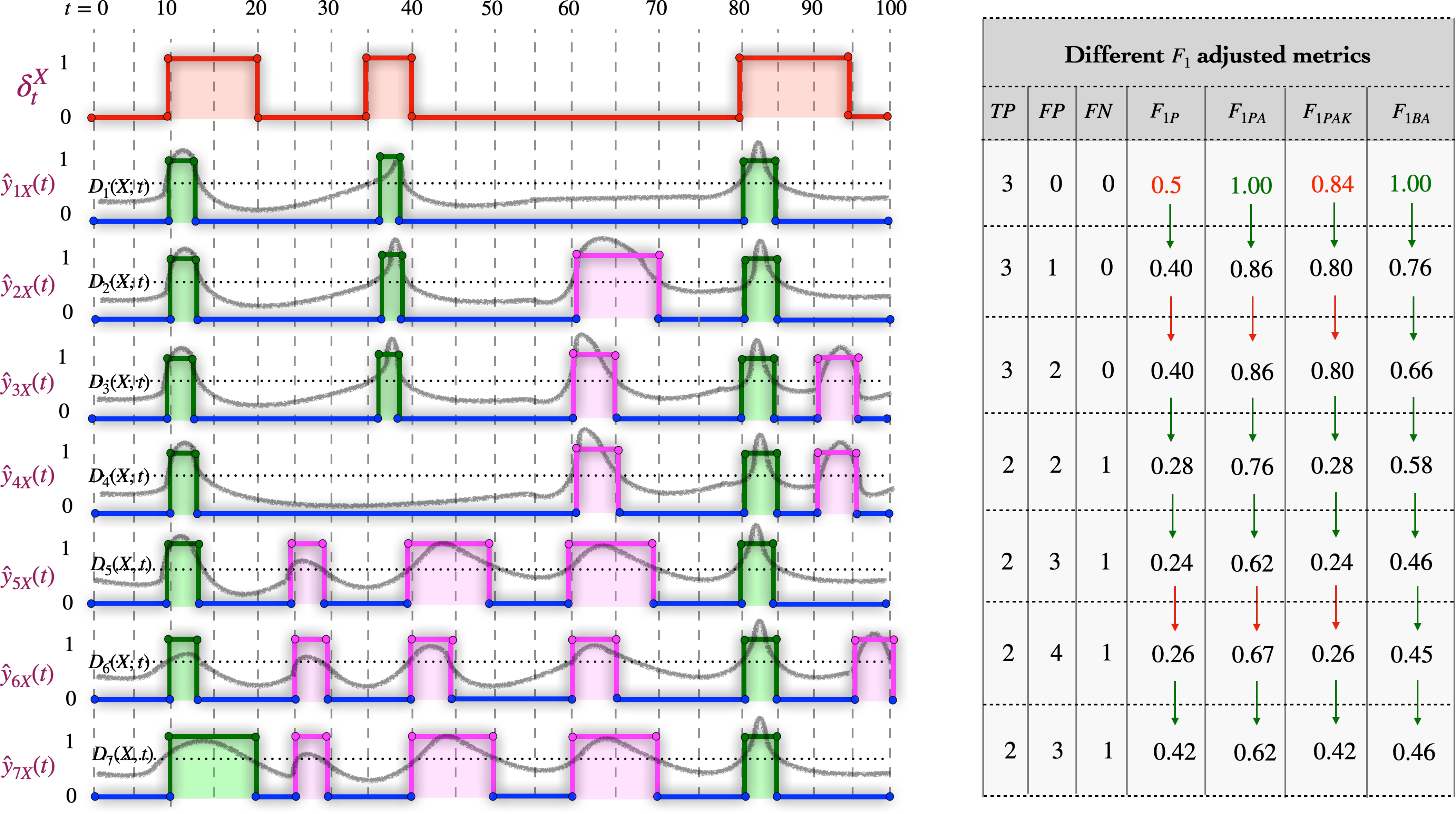} \caption{Comparison of $F_{1p}$, $F_{1PA}$, $F_{1KPA}$, and our proposed $F_{1BA}$. In the table, the green color shows ideal metric values for a perfect detection, while the red color highlights failure to indicate correct predictions. The proposed $F_{1BA}$ consistently makes meaningful transitions, unlike other metrics.} \vspace{-0.2in} \label{fig:fair-comparison} 
\end{figure}

%% file: sections/methods.tex
\section{Methods}
\subsection{Notations}
\subsubsection{Time-series}
Let's consider univariate time-series sample space $\mathcal{X}$. Time series $X$ of length $T$ can be sampled from $\mathcal{X}$: $X \in \mathcal{X}$, so that $X = (x_1,x_2,\dots,x_T)$
\subsubsection{Anomaly labels}
For a given time series $X$, anomaly labels is a time series $\delta^{X} = (\delta^{X}_1, \delta^{X}_2, \cdots, \delta^{X}_{T})$ with $\delta^{X}_t \in \{0,1\} \; \forall t$.
\subsubsection{Anomaly segment}
An anomaly segment is a contiguous subsequence of timesteps corresponding to an anomaly event. We define an $i^{th}$ anomaly segment occurring in $X$ as:
\begin{equation}
    \mathcal{A}^{X}_{i}:=(a_{i}, w_{i}, s_{i})
\end{equation}
where $a_{i}$, $w_{i}$ and $s_{i}$ denote the starting timestamp, time-width, and severity of the $i^{th}$ anomaly segment respectively. Note that,
\\
$\delta_{t}^X=\begin{cases}
1 & t \in \cup_{i} \{a_{i}, a_{i+1}\cdots a_{i+w_{i}-1}\} = \cup_{i} S_a^i\\
0 & \text{otherwise}
\end{cases}$

\noindent where $S_a^i$ denotes the set of time-steps corresponding to an anomaly event, $\{a_{i}, a_{i+1}\cdots a_{i+w_{i}-1}\}$.
\subsubsection{Anomaly detector}
An anomaly detector $\mathcal{D}(t, X)$ labels a timestep $t$ in $X$ as an anomaly point if
\begin{equation}
    D(t, X) > \gamma
\label{eqn:detector-thresholding}
\end{equation}
with $t\in [0, T-1]$ and $\gamma$ being a threshold.
\subsubsection{Metric for time-series anomaly detection}
Let the metric to quantify the anomaly detection performance of an anomaly detector $\mathcal{D}(\cdot, \cdot)$ is $\mathcal{M}(y, \delta)$, where $y(t) = D(t, X)$, $\delta\in \{0,1\}^T, y \in \mathbb{R}^T$.

\subsection{Illustrative analysis}
We provide analysis of the different metrics concerning diverse scenarios with respect to True Positives (TP),  False Positives (FP), False Negatives (FN) numbers in Figure~\ref{fig:fair-comparison}. The table in the right panel of Figure~\ref{fig:fair-comparison} studies different metric values for diverse scenarios of TP, FP, and FN events. Firstly, $F_{1p}$ and $F_{1KPA}$ fail to indicate the perfect detection $\hat{y}_{1X}(t)$ by detector $D_1$. Further, with varying number of TP, FP, and FN events, the expected transitions (increase/decrease) of different metrics are also studied. We observe, that the proposed metric $F_{1BA}$ makes consistent transitions with excellent coverage.

\subsection{Axiomatic criterion for TSAD metrics}
A time-series anomaly detector requires a robust and reliable metric for accurate comparison with other detectors. Based on existing literature, we formalize the essential requirements of a TSAD metric using the following axioms: (a) The metric should be resistant to random noise and uncorrelated data~\citep{pa-ka}, (b) it should reward better detectors with higher scores, and (c) it should grant the best score exclusively to the best detection performance. To enable use of  standard mathematical tools, we analyze the detector scores $y = D(X, t)$ directly, rather than binary predictions derived from thresholding (Eqn.~\ref{eqn:detector-thresholding}) as defined formally below.\\
A TSAD metric $\mathcal{M}(y, \delta)$ is said to be a \textit{good} metric if it satisfies the following properties:
\begin{itemize}
    \item \textbf{\textit{C-1 (robust)}}: For any random detection signal $y_{random}$ ($\perp\!\!\!\perp \delta$), the metric value should always be less than its chance level value ($\mathcal{M}_{chance} = 0.5$ for $F_1$ score based metrics), 
    \begin{equation}
        \mathcal{M}(y_{random}, \delta) \leq \mathcal{M}_{chance}
    \end{equation}
    almost surely.
    \item \textbf{\textit{C-1a (threshold agnostic)}}: If $y_{random}\sim \mathcal{U}[0, 1]$ uniformly distributed noise, then $\mathcal{M}(y_{random}, \delta)$ should remain unaffected by threshold variation.
    \begin{equation}
        \mathcal{M}(y_{random}, \delta) \rightarrow \mathcal{M}_u, \, \forall \, \gamma
    \end{equation}
    Note that $\mathcal{M}_u$ is a constant ($\mathcal{M}_u \leq \mathcal{M}_{chance}$).
    \item \textbf{\textit{C-2 (ordered)}}: For two detectors $D_1(\cdot, \cdot)$ and $D_2(\cdot, \cdot)$ with discriminability order $D_2 \geq D_1$, the following holds,
    \begin{equation}
        \mathcal{M}(y_1, \delta) \geq \mathcal{M}(y_2, \delta) \label{eq:order-axiom}
    \end{equation}
    where $y_1 = D_1(t, X)$ and $y_2 = D_2(t, X)$.
    \item \textbf{\textit{C-3 (exclusive)}}: The metric should reach maximum value for a perfect detection signal ($y_{prf}$) with no FP and FN events ($\mathcal{M}(y_{prf}, \delta) \rightarrow \mathcal{M}_{max}$), and should fail to converge if a single FP/FN occurs.
    \begin{equation}
        \mathcal{M}(y_{prf} + \{\text{1 FP/FN}\}) \rightarrow \mathcal{M}_{max} - \epsilon, \epsilon > 0
    \end{equation}
\end{itemize}
\vspace{-0.1cm}

\subsection{Metric analysis}
\subsubsection{\textbf{Pre-requisites}}
To allow analysis of the proposed metric and existing ones, we first need their closed-form expressions. We find out the limiting expressions of $F_{1PA}$ and $F_{1BA}$ first.
\begin{definition}[\textbf{Point adjustment}]
The point adjustment procedure involves the following steps:
\begin{itemize}
    \item \textbf{Thresholding}: 
    \begin{equation}
        \hat{y}_{X}(t) = 
    \begin{cases}
        1 & \text{if}\, D(X,t) > \gamma \\
        0 & \text{if}\, D(X,t) \leq \gamma
    \end{cases}
    \label{eqn:thresholding_PA}
    \end{equation}
    \item \textbf{Point-adjustment (PA)}: 
    \begin{equation}
        \hat{y}_{X, PA}(t) =
    \begin{cases}
    1 & \text{if}\, \hat{y}_{X}(t) = 1 \\
        1 & \text{if}\, t\in S_a^i \,\text{and}\, \exists \,t'\in S_a^i \,\text{s.t.}\, \hat{y}_{X}(t') = 1 \\
        0 & \text{elsewhere}
    \end{cases}
    \label{eqn:point_adjustment_PA}
    \end{equation}
    
    \item 
    \textbf{$\mathbf{F_{1PA}}$ score}:
    \begin{equation}
        F_{1PA} = F_1(\text{true} = \delta, \text{prediction} = \hat{y}_{X, PA}(t))
        \label{eqn:f1_score_PA}
    \end{equation}
\end{itemize}
\end{definition}
In contrary, our proposed Balanced Adjustment (BA) procedure can be detailed as:
 \begin{definition}[\textbf{Balanced Adjustment (BA)}] The BA involves:
\begin{itemize}
    \item \textbf{Thresholding}: Based on Eqn.~\ref{eqn:thresholding_PA}.
    
    \item \textbf{BA}: \\
    We define islands of width $w_N$ at time $u$ around FPs as:
    \begin{multline}
        S_N(u) := \{k : u-\frac{w_N}{2}-1 \leq k \leq u+\frac{w_N}{2},\,\\ \hat{y}_X(u) = 1,\, \delta(u) = 0\}
    \end{multline}
    Clearly, $|S_N(u)| = w_N, \forall u$ and it is defined only at FP time steps. Using the islands $S_N(u)$, we define adjusted prediction:
    \begin{equation}
        \hat{y}_{X, BA}(t) = 
        \begin{cases}
            1 & \text{if}\, \hat{y}_{X}(t) = 1 \\
        1 & \text{if}\, t\in S_a^i \,\text{and}\, \exists \,t'\in S_a^i \,\text{s.t.}\, \hat{y}_{X}(t') = 1 \\
        1 & \text{if}\, \exists\, u,\,  t\in S_N(u)\\
        0 & \text{elsewhere}
        \end{cases}
    \end{equation}
    \item \textbf{$\mathbf{F_{1BA}}$ score}:
    \begin{equation}
        F_{1BA} = F_1(\text{true} = \delta, \text{prediction} = \hat{y}_{X, BA}(t))
    \end{equation}
\end{itemize}
\end{definition}
\noindent Now, we obtain expressions of $F_{1PA}$ and $F_{1BA}$ based on definitions.
\begin{theorem}[$\mathbf{F_{1PA}}$ \textbf{in random noise}]
\label{thm:f1_pa_formulation}
\textit{
The point-adjusted (PA) F1 score ($F_{1PA}$) of any random time-series anomaly detector working on a sufficiently large time series of length $T$ having a single anomaly event ($S_A := S_a$) is:
\begin{align}
    & F_{1PA}  =  \frac{2q \left(1 - N(\gamma)^{|S_a|}\right)}{\left(1 - N(\gamma)\right) + q \left(1 + N(\gamma) - N(\gamma)^{|S_a|}\right)}
\label{eqn:PA_formula}
\end{align}
where $q = \frac{|S_a|}{T}$ is the anomaly ratio, $N(\cdot)$ is the noise cdf.
}
\end{theorem}
\vspace{-0.2cm}
\begin{proof}
    For sufficiently long time-series ($T\rightarrow \infty$), 
\begin{eqnarray}
     R_{PA}  &=& Pr\left(\hat{y}_{X, PA}(t) = 1 \,|\, \delta(t) = 1\right)\nonumber\\
    & = & 1 - \prod_{t\in S_a}Pr\left(\hat{y}_{X}(t) \leq \gamma\right)\nonumber\\
    & &(\because \hat{y}_{X}(t_i)\perp\!\!\!\perp \hat{y}_{X}(t_j), i\neq j\nonumber \, \text{as}\, \hat{y}_{X} \sim N \text{(noise)})\nonumber\\
    & = & \left(1 - N(\gamma)^{|S_a|}\right)
\label{eqn:r_pa}
\end{eqnarray}
\begin{eqnarray}
     P_{PA}  &=& Pr\left(\delta(t) = 1 \,|\, \hat{y}_{X, PA}(t) = 1\right)\nonumber\\
     & = & \frac{Pr\left(\hat{y}_{X, PA}(t) = 1 \,|\, \delta(t) = 1\right) Pr\left(\delta(t) = 1\right)}{Pr\left(\hat{y}_{X, PA}(t) = 1\right)}\nonumber\\
     & = & \frac{q \left(1 - N(\gamma)^{|S_a|}\right)}{\left(1 - N(\gamma)\right) + q(N(\gamma) - N(\gamma)^{|S_a|})}
\label{eqn:p_pa}
\end{eqnarray}
The expression of $F_{1PA}$ follows.
\end{proof}
\begin{figure}[t!]
  \centering
  \includegraphics[width=\linewidth]{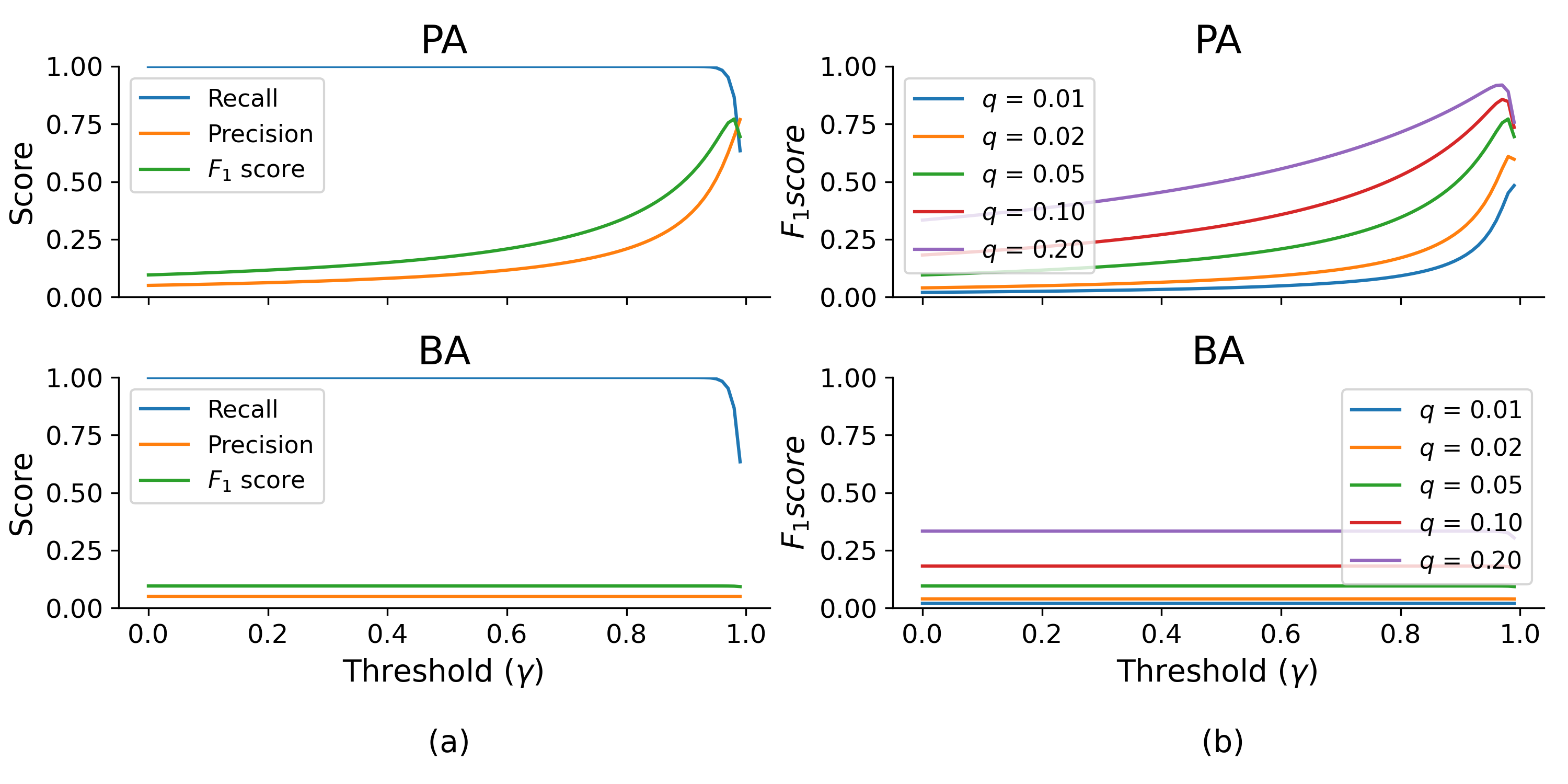}
\caption{(a) The behavior of BA metrics $P_{BA}, R_{BA}, F_{1BA}$ compared to PA metrics for scores from uniform noise with varying thresholds $\gamma$, using anomaly width of $100$ and ratio $q=0.2$. $F_{1PA}$ rises above $0.75$ for random anomaly scores, (b) The right panel illustrates the behavior of $F_{1PA}$ and $F_{1BA}$ with varying $\gamma$ for different anomaly ratios ($q$), with anomaly width of 100. $F_{1PA}$ increases with higher thresholds, while $F_{1BA}$ remains unaffected by threshold choice.}
  \vspace{-0.2in}
  \label{fig:pa_vs_BA_in_noise}
\end{figure}
\begin{theorem}[$\mathbf{F_{1BA}}$ \textbf{in random noise}]
\label{thm:f1_BA_formulation}
\textit{
The balanced point-adjusted (BA) F1 score ($F_{1BA}$) of any random time-series anomaly detector working on a sufficiently large time series of length $T$ having a single anomaly event ($S_A := S_a$) is:
\begin{align}
    & F_{1BA}  =  \frac{2q \left(1 - N(\gamma)^{|S_a|}\right)}{\left(1 - N(\gamma)^{w_N}\right) + q \left(1 + N(\gamma)^{w_N} - N(\gamma)^{|S_a|}\right)}
\end{align}
}
\end{theorem}
\begin{proof}
\vspace{-0.1in}
Assume that the minimum separation between false positive predictions in $\hat{y}_X(t)$ is more than the island width ($w_N$).
    Now, $R_{PA}$ remains unaltered as in PA.
    \begin{eqnarray}
     P_{BA}  &=& Pr\left(\delta(t) = 1 \,|\, \hat{y}_{X, BA}(t) = 1\right)\nonumber\\
     & = & \frac{q(1 - N(\gamma)^{|S_a|})}{(1 - N(\gamma)^{w_N})+q(N(\gamma)^{w_N} - N(\gamma)^{|S_a|})}
\label{eqn:p_BA}
\end{eqnarray}
    The expression of $F_{1BA}$ follows. Interestingly, expressions hold same structure as in Eqn.~\ref{eqn:PA_formula} with additional exponentials of $w_N$.
\end{proof}
The Figure~\ref{fig:pa_vs_BA_in_noise} studies the $F_{1PA}$ and $F_{1BA}$ for uniformly randomly drawn noise as anomaly score. It shows that $F_{1PA}$ increases by threshold selection and can give a very high score as well. However, $F_{1BA}$ score is unaffected by threshold choice and remains below $0.5$. \\
\subsubsection{\textbf{\textit{C-1 (robust)}}}
\begin{lemma}\label{lemma:randomness}
For $w_N = |S_a|$, the $F_{1BA}$ is always less than equal to the chance level $F1$ score of $0.5$ for any randomly generated anomaly score, as long as the anomaly ratio ($q$) is less than equal to $\frac{1}{3}$.
\begin{equation}
    F_{1BA}(\delta, \hat{y}_{X, BA}(t)) \leq F1_{chance} = 0.5, \, X\sim N(\cdot) \, \forall N(\cdot)
\end{equation}
\end{lemma}
\begin{proof}
Assuming the anomaly event is sufficiently sustained (not momentary) so that the anomaly width ($|S_a|$) is not very small,
\begin{eqnarray}
     F_{1BA}  
     & = & \frac{2q(1 - N(\gamma)^{|S_a|})}{(1 - N(\gamma)^{|S_a|}) + q} \, (\text{using $|S_a| = w_N$})\nonumber\\
     & = & \frac{2q}{1 + q} (\because |S_a|>>1 \implies N(\gamma)^{|S_a|} \rightarrow 0)
\label{eqn:f1_less_than_chance}
\end{eqnarray}
Now, $\frac{2q}{1+q} \leq 0.5 \implies q\leq \frac{1}{3} = 33.33\%$
\end{proof}
\subsubsection{\textbf{\textit{C-1a (threshold agnostic)}}}
\begin{lemma}
If an anomaly scorer generates random score from uniform distribution, the $F_{1BA}$ not only stays less than the chance value, but also behaves as threshold agnostic (stays constant across all possible thresholds $\gamma$) for almost the entire range of thresholds as long as the anomaly width is not very small. 
\begin{eqnarray}
    F_{1BA}(\delta, \hat{y}_{X, BA}(t)) \leq  F1_{chance} = 0.5, \, X\sim \mathcal{U}[0, 1] \nonumber\\
    \frac{\partial F_{1BA}(\delta, \hat{y}_{X, BA}(t))}{\partial \gamma} \rightarrow 0, \forall \gamma \in (\gamma_{min}, \gamma_{max})&&
\end{eqnarray}
\end{lemma}
\begin{proof}
Because lemma~\ref{lemma:randomness} hold for all $N(\cdot)$, it holds for $\mathcal{U}[0,1]$ too. Now, using \text{$ N(\gamma) = \gamma \,\text{for}\, N(\cdot) := \mathcal{U}[0,1]$},
\begin{eqnarray}
\frac{\partial F_{1BA}}{\partial \gamma} 
  & = & - \frac{2q^2|S_a|\gamma^{|S_a|-1}}{(1 - \gamma^{|S_a|} + q)^2}
\end{eqnarray}
Note that, $F_{1BA}$ is monotonic w.r.t. $\gamma$: $\frac{\partial F_{1BA}}{\partial \gamma} \leq 0$.
Further, 
\begin{eqnarray}
\frac{\partial F_{1BA}}{\partial \gamma} &=&  - 2 . \left(\frac{q}{(1 + q - \gamma^{|S_a|})}\right)^2 . |S_a|\gamma^{|S_a| - 1}\nonumber
\end{eqnarray}
with 
\begin{equation}
    \frac{q}{(1 + q - \gamma^{|S_a|})} < 1 \,\text{as}\, |q| < 1,\, |S_a| >> 1
\end{equation}
\begin{equation}
    |S_a|\gamma^{|S_a| - 1} \rightarrow 0 \,\text{as}\,  |S_a| >> 1
\end{equation}
It can be formally shown using limit: 
\begin{equation}
    \mathop{lim}_{x\rightarrow \infty}x.b^{x-1} = 0,\, 0<b<1
\end{equation}
Hence, $\frac{\partial F_{1BA}}{\partial \gamma} \rightarrow 0$, implying $F_{1BA}$ remains constant across $\gamma$.
\end{proof}
\subsubsection{\textbf{\textit{C-2 (ordered)}}}
\begin{lemma}
Let, for any detector $D(t, X)$ has the following parameters:
\begin{align}
    & \alpha = Pr(\hat{y}_X(t) \leq \gamma | \delta(t) = 1)\nonumber\\
    & \beta = Pr(\hat{y}_X(t) \leq \gamma | \delta(t) = 0)
\end{align}
Then, for two different detectors $D_1$ and $D_2$ with strength order ($D_1>D_2$) given by the conditions: $\alpha_{D_1} \leq \alpha_{D_2}, \, \beta_{D_1} \leq \beta_{D_2}$
, the following holds,
\begin{align}
    & F_{1BA}\left(\hat{y}_{X, BA}(t), \delta(t)\right) \leq F_{1BA}\left(\hat{z}_{X, BA}(t), \delta(t)\right)\nonumber\\
    & \hat{y}_{X}(t) = D_1(t, X), \, \hat{z}_{X}(t) = D_2(t, X)
\end{align}
\end{lemma}
\begin{proof}
For Detector $D_i$ ($i\in\{1,2\}$):
\begin{align}
    & F_{1BA}^{D_i} = \frac{2q(1 - \alpha_{Di}^{|S_a|})}{1 - \beta_{Di}^{|S_a|} + q(1 + \beta_{Di}^{|S_a|} - \alpha_{Di}^{|S_a|})}
\end{align}
So, $F_{1BA}(\alpha_1, \beta) \geq F_{1BA}(\alpha, \beta) \forall \alpha>\alpha_1$ and $F_{1BA}(\alpha, \beta) \geq F_{1BA}(\alpha, \beta_2) \forall \beta_2>\beta$ gives $F_{1BA}(\alpha_1, \beta_1) \geq F_{1BA}(\alpha_2, \beta_2)$.
\end{proof}
\subsubsection{\textbf{\textit{C-3 (exclusive)}}}
\begin{lemma}
For a perfect anomaly signal $y_{prf}$ that detects the anomaly event correctly and gives no false alarm, and $y_1 = y_{prf} + f_{k}$ being the score which triggers $k$ false alarms,
\begin{equation}
    F_{1BA}(\delta, \hat{y}_{prf, BA}(t)) = 1
\end{equation}
\begin{equation}
    F_{1BA}(\delta, \hat{y}_{1, BA}(t)) = 1 - \epsilon, \, \epsilon > 0
\end{equation}
\end{lemma}
\begin{proof}
By definition, in absence of any FP/FN in $\hat{y}_{prf, BA}(t)$, $F_{1BA}(\delta, \hat{y}_{prf, BA}(t)) = 1$. Now, for a tiny FP event of width $w_{fp}<< |S_a|$, $P_{PA} = \frac{|S_a|}{|S_a| + w_{fp}}\rightarrow 1$, $P_{BA} = \frac{|S_a|}{|S_a| + w_{fp} + w_N} \rightarrow 1-\epsilon, \, \epsilon>0$. As $R_{PA} = R_{BA} = 1$, $F_{1BA} \rightarrow 1-\epsilon, \, \epsilon>0$. Note, $F_{1PA} \rightarrow 1$.
\end{proof}

%% file: sections/experiments.tex
\section{Experiments}


This section provides a detailed empirical evaluation of $F_{1BA}$ against other F1 scores through controlled experiments. We have implemented a controlled experimental setup that allows us to generate anomaly label sequences and detector predictions with a control on the number of anomaly events, TP, FP, and detector score quality. The setup enables the empirical study of the scaling behavior of F1 scores with other metrics.

\subsection{Data preparation}
We develop a simulation tool for data preparation. The simulator generates the controlled anomaly sequences in two steps. First, it produces the ground truth label ($\delta^X$), with constraints on the total number of anomalies, the width of an anomaly event, and the separation between two successive events. The second step yields a controlled generation of the detector scores $y = D(t, X)$. The second step consists of three sub-steps, (1) latent detection, (2) score simulation, and (3) threshold-driven anomaly marking.

Many measurements are computed from simulated labels, along with \emph{F1 scores}. 
Let $N$ and $M$ represent the number of anomaly events in the ground truth and the predictions respectively. $m^{s}_{tp}$ denotes the number of true events that are detected, and $n^{s}_{tp}$ is the number of true positives in prediction. It must be noted that $m^{s}_{tp} \leq n^{s}_{tp}$, as a single anomaly event can overlap with multiple detections. $c_{i}$ is the anomaly detection strength of $i^{th}$ event.


We derive the essential attributes from the simulated samples that are important for our study of $F_1$ metric scores.
\paragraph{\bf Separation Score}
\label{def-sep}
It is defined as the distribution distance (Hellinger distance~\citep{van2000asymptotic}) between the regular and anomalous scores. $H(P, Q) = \frac{1}{\sqrt{2}}\sqrt{\sum^{k}_{i=1}{(\sqrt{p_i} - \sqrt{q_i})}^{2}}$, where $P$ and $Q$ are discrete probability distribution over regular and anomalous region scores. 
$P$ and $Q$ are evaluated only over the overlapped score values, using non-parametric kernel density estimates. A higher value signifies a better anomaly detector. 

\paragraph{\bf Precision}
\label{def-prec}
We define the precision metric at an event label as Precision${}_{E}$ = $\frac{m^{s}_{tp}}{M}$. The precision uses predicted labels only.

\paragraph{\bf Recall}
\label{def-recall}
Similar to precision, recall is computed at an event level with the ground-truth label, Recall${}_{E}$ = $\frac{n^{s}_{tp}}{N}$.

\paragraph{\bf Coverage}
\label{def-cov}
This metric measures the average fraction of the detected true events, $\frac{1}{n^{s}_{tp}}\sum^{N}_{i=1} c_{i}$.

We have conducted $15,000$ distinct controlled simulated experiments with varying TP, FP, and total anomaly count. We uniformly sample coverage, and separation score, using parameter grid search. 
$F_{1KPA}$ computation uses $K=20\%$. 

\subsection{Study of metric behavior}

We study the scaling relation of $F_{1}$ scores with aforementioned $4$ most important anomaly detector characteristics- \textit{precision}, \textit{recall}, \textit{detection coverage}, and  \textit{separation score}.
\subsubsection{F1 scores vs separation score with varying recall}
\begin{figure}[htp!]
    \centering
    \includegraphics[width=\linewidth]{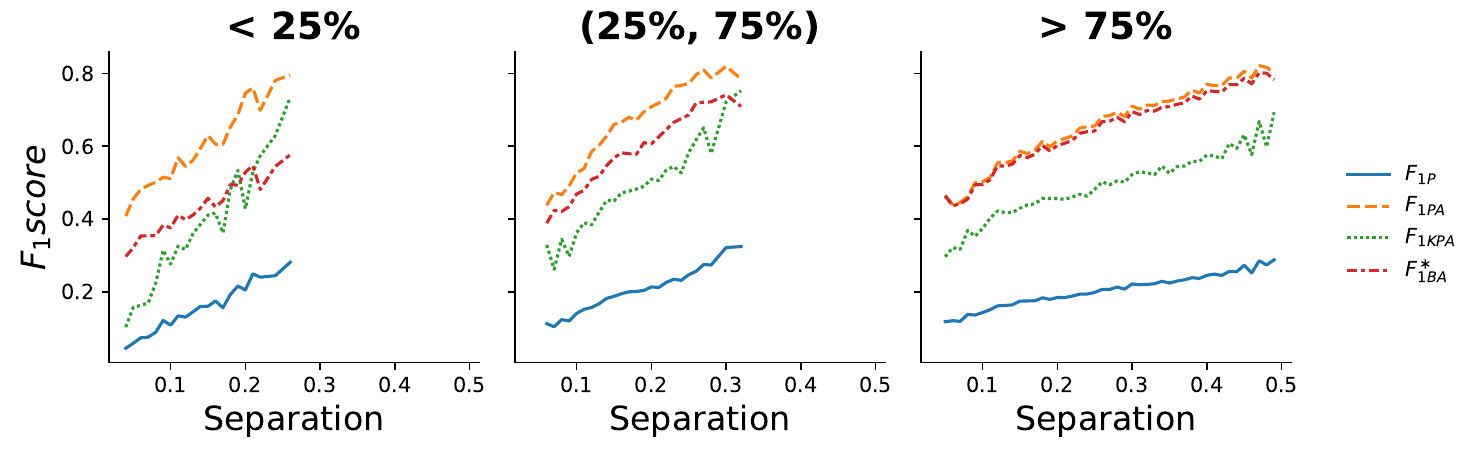}
    \caption{Metric behavior plotted against the score separation metric (\ref{def-sep}). Plots are made for varying recall~\ref{def-recall} in $3$ different bins of $< 25\%, (25\% - 75\%)$ and $>75\%$. The bins are chosen so that similar data point cardinality is maintained. Precision is maintained within $(25\% - 75\%)$.
    }
    \vspace{-0.1in}
    \label{fig:sep-sens}
\end{figure}

The behaviour of different metrics against separation score is shown in Figure~\ref{fig:sep-sens} for three different recall ranges, separately. Firstly, all the $4$ metrics show an increasing trend with separation. 
The increase in the separation score suggests better anomaly identification with a constant threshold, which results in an increasing F1 value.  
However, we note the following three observations: (a) The $F_{1P}$ is always very low, even for a high recall and high separation, (b) for low recall of $<25\%$ (and precision maintained between $25\% - 75\%$), $F_{1PA}$ and $F_{1KPA}$ achieve very high score close to $0.8$, which indicates biased behaviour of the metrics, while $F_{1BA}$ is least affected as it penalizes false positives. (c) for high recall, $F_{1BA}$ and $F_{1PA}$ converges, indicating applicability of $F_{1PA}$ at only high recall setting.
\subsubsection{F1 scores vs precision with varying coverage}
\begin{figure}[htp!]
\centering
    \includegraphics[width=\linewidth]{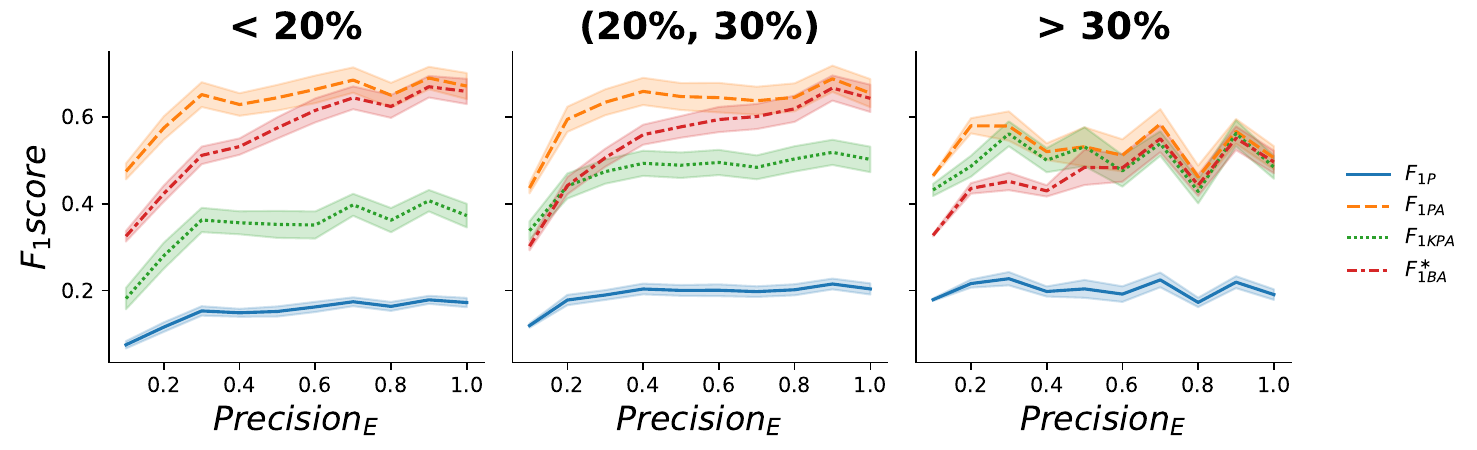}
    \caption{
    Metric behavior plotted against the precision (\ref{def-prec}). Plots are made for varying coverage score~\ref{def-cov} in $3$ different bins of $< 20\%, (20\% - 30\%)$ and $>30\%$. The bins are chosen so that similar data point cardinality is maintained. Recall is maintained within $(25\% - 75\%)$.
    }
    \vspace{-0.1in}
    \label{fig:prec-sens}
\end{figure}

The behaviour of different F1 scores against precision is shown in Figure~\ref{fig:prec-sens} for three different ranges of coverage, separately with medium recall value $(25\% - 75\%)$. We make the following remarks: (a) Among all the metrics, the $F_{1BA}$ shows the highest sensitivity (high slope) to precision which shows the stricter compliance to the \textit{ordering} axiom (equation~\ref{eq:order-axiom}), (b) The $F_{1PA}$ is an overestimate at the low precision and the $F_{KPA}$ is an underestimate at high precision. $F_{1BA}$ behaves like $F_{1KPA}$ at low precision and like $F_{1PA}$ at high precision.
Note that all the metrics show drop in score for high coverage and high precision as we maintain medium recall while generating the samples. 
\subsubsection{F1 scores vs recall with varying coverage}
\begin{figure}[htp!]
\centering
    \includegraphics[width=\linewidth]{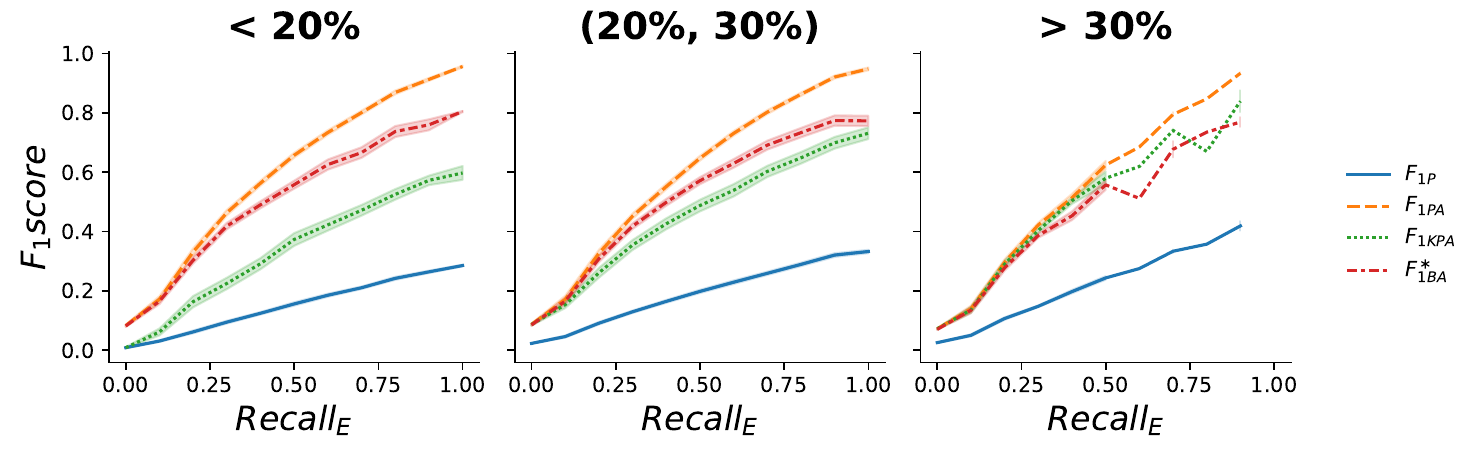}
    \caption{
    Metric behavior plotted against the recall (\ref{def-recall}). Plots are made for varying coverage score~\ref{def-cov} in $3$ different bins of $< 20\%, (20\% - 30\%)$ and $>30\%$. The bins are chosen so that similar data point cardinality is maintained. Precision is maintained within $(25\% - 75\%)$.
    }
    \vspace{-0.1in}
    \label{fig:recall-sens}
\end{figure}

The behaviour of different metrics against recall is shown in Figure~\ref{fig:recall-sens} for three different ranges of coverage, separately with medium precision value $(25\% - 75\%)$. We note the following: (a) Because of the maintained precision range, the metric scores should not cross $0.75$ even for the highest recall, as the $F_1$ score is a harmonic mean of precision and recall. However, $F_{1PA}$ approaches $1.0$ for all coverage values. Similar to F1 scores vs separation score case (Figure~\ref{fig:sep-sens}), this arises from inappropriate penalization of false positives, overestimating the precision, (b) $F_{1KPA}$ and $F_{1P}$ continue to behave as underestimate.\\


To summarize, the experiments show $F_{1BA}$ follows the axiomatic criteria better than the other metrics.

%% file: sections/conclusions.tex
\section{Conclusion \& Future Work}  
In this work, we introduced an axiomatic approach for ordering of anomaly detection models. We proposed a new $F_1$ score ($F_{1BA}$) and motivated it with an illustrative example. We provided theoretical proofs of the compliance of the proposed score to the axioms. Additionally, we developed a simulation setup for controlled metric comparison. Detailed experimental results demonstrate the efficacy of our proposed score. \\
We believe this contribution will aid both the applied and research communities by facilitating a more systematic and unbiased approach to anomaly model selection.